\documentclass[journal]{IEEEtran}
\usepackage{cite}
\usepackage{multirow}
\usepackage{color}
\usepackage{graphicx}
\usepackage{subfig}
 \usepackage{amsmath}
\usepackage{amssymb}
\usepackage{floatrow}
\usepackage{booktabs}

\usepackage{algorithm}  
\usepackage{algorithmic}

\usepackage[colorlinks,linkcolor=red,citecolor=green,urlcolor=blue]{hyperref}

\renewcommand\thesection{\Roman{section}}

\ifCLASSINFOpdf
\else
\fi
\begin{document}
\title{Content-adaptive Representation Learning for Fast Image Super-resolution}

\author{Yukai Shi~\textsuperscript{*} , Jinhui Qin~\textsuperscript{*} 
~\thanks{

* The first two authors share first-authorship.
}
}
\markboth{}%
{Shell \MakeLowercase{\textit{et al.}}: Bare Demo of IEEEtran.cls for IEEE Journals}

\maketitle

\begin{abstract}
Deep convolutional networks have attracted great attention in image restoration and enhancement. Generally, restoration quality has been improved by building more and more convolutional block. However, these methods mostly learn a specific model to handle all images and ignore difficulty diversity. In other words, an area in the image with high frequency tend to lose more information during compressing while an area with low frequency tends to lose less. In this article, we adrress the efficiency issue in image SR by incorporating a patch-wise rolling network(PRN) to content-adaptively recover images according to difficulty levels. In contrast to existing studies that ignore difficulty diversity, we adopt different stage of a neural network to perform image restoration. In addition, we propose a rolling strategy that utilizes the parameters of each stage more flexible. Extensive experiments demonstrate that our model not only shows a significant acceleration but also maintain state-of-the-art performance.

\end{abstract}


\section{Introduction}
\label{}
Deep learning has successfully applied in many computer vision fields such as image recognition~\cite{residual_net}, semantic segmentation~\cite{unet} and object detection~\cite{ouyang2015deepid}. Inspired by the rapid development and superior performance, many efforts have been made to introduce deep learning in low-level vision as well as image processing tasks, including image suer-resolution~\cite{srcnn}, image enhancement~\cite{dped}, inpainting~\cite{shepard} etc. Meanwhile, Single image super-resolution(SISR), namely to predict high-resolution with low-resolution input, is widely used in many computer vision applications and draws plenty of attentions~\cite{srcnn,vdsr,srgan,edsr,subpixel,laplacian,fsrcnn}.

Recently, Convolutional neural networks(CNNs) achieve magnificent improvement toward image restoration by adopting a building block strategy. VDSR~\cite{vdsr} utilizes residual connection and a very deep model to achieve promising results in image SR. EDSR~\cite{edsr} further improves the results by adopting residual block~\cite{residual_net} and remove batch normalization. However, they advance performance with numerous parameter gain and huge computational cost. Dense block~\cite{densenet} also exhibits its effectiveness in image enhancement. MemNet~\cite{memNet} realizes a coarse-to-fine restoration process by using dense block and recursive unit. Zhang \emph{et al.}~\cite{residualdensenet} proposes an optimized block, which combines the strengths of the dense block and residual block, and achieve impressive promotion. However, deep learning-based SR methods~\cite{vdsr,memNet,residualdensenet,edsr} prefer to crop the image into patches before training phrase. As different patch has various texture and structure, it is inefficient to adopt a feed-forward network to super-resolve all samples, especially for those \emph{intensely simple} patches. In addition, notwithstanding such a complicated model can bring positive performance with a graphics processing unit(GPU), it also leads to expensive computational cost and explosion of parameters.

\begin{figure}[t]
\centering
\includegraphics[width=0.88\textwidth]{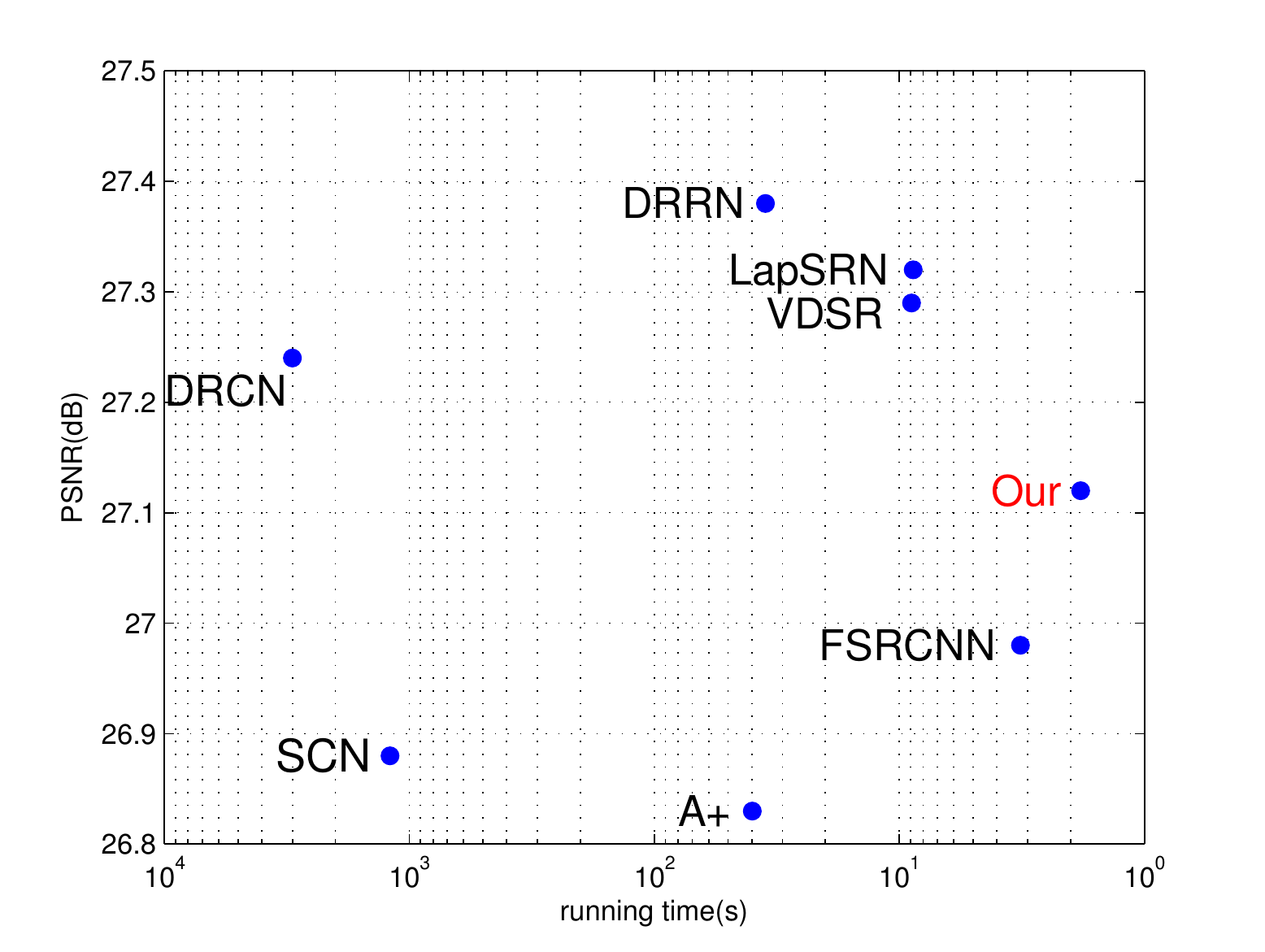}
\caption{Efficiency analysis of the proposed model. The results are evaluated on BSDS100 with factor 4$\times$. The proposed model runs twice as fast as state-of-the-art SR inferences with superior performance. }
\label{fig:efficiency}
\end{figure}

Computer vision applications and technongies~\cite{dped,shufflenet,mobilenet} for mobile devices draw a lot of attention as it has wide application scenarios. However, using CNNs on the mobile platform has an extreme requirement towards efficiency. MobileNet~\cite{mobilenet} makes an attempt to accelerate speed by utilizing a depth-wise convolution to reduce redundancy of CNNs. Similar technology also adopted by ShuffleNet~\cite{shufflenet}. Moreover, ShuffleNet employs a novel shuffle unit, which maintains performance with efficiency improvement. However, their methods are limited by the optimization of the computational platform and sometimes run inefficiently. IGC~\cite{IGC} utilizes parameters of the deep network more efficiently by adopting group convolution and permutation of convolutional features. The similar idea also used by RRC~\cite{RRC}. RRC implements a rolling strategy on object detection, which not only utilizes multi-scale features but also realizes an efficient one-stage framework. Their methods reveal that features of different scale can be utilized more efficiently. MSDNet~\cite{msdnet} proposes a multi-scale dense net, which adaptively uses the specific stage in the deep model to deal with samples with different difficulty levels. For instance, MSDNet adopts early stage convolutional layers to handle easy samples and more parameters are applied to process difficult images. However, MSDNet can inherently distinguish difficult level with an internal high-level representation of the image itself. Since such internal high-level prior is not exist in low-level vision, MSDNet is fail to applied in images processing tasks.
 
Motivated by previous works, we make an attempt to propose a content-adaptive and flexible framework, which can accurately super-resolve image with different difficulty level according to gradient prior. In the proposed model, we first define the gradient prior to distinguish different samples. Then, a unified model is proposed to handle samples with different difficulty by a content-adaptive fashion. Since samples with different difficulty will cause frequency conflicts and result in a performance degradation. We also propose a flexible rolling strategy by alternating the convolution filters to address this problem. 

Our main \textbf{contributions} are summarized as follows.
\begin{itemize}
\item We find it is inefficient to adopt an expensive model to mild samples, which have less texture and simple structure. In contrast, an expensive model is appropriate for the samples, which have rich texture and complicated structure.

\item According above observation, we distinguish the difficulty of samples by its gradient prior and content-adaptively adopt different convolutional stage to super-resolve samples. This strategy helps us greatly improve SR efficiency.

\item Since the samples with different difficulty exhibit various property in the frequency domain, which causes frequency conflicts and leads to a performance degradation. We propose a flexible rolling strategy. With our rolling approach, our model not only achieve a balance between mild and severe samples but also increase the receptive field of early layers.
\end{itemize}

\section{Related Work}

\textbf{CNN for image SR.} 
Recently, deep learning based SR methods have achieved a great successes in many computer vision fields. Super-resolution, which considered a typical low-level vision task and is well-known for its ill-posed property, plays an important role in image quality enhancement. Many researchers devote themselves to the studies of super-resolution and have proposed many insightful works. Recently, the rising of deep learning methods give new solution to image SR. Dong~\emph{et al.}~\cite{srcnn} first adopt deep convolutional neural networks to learn the mapping from LR to HR patches in an end-to-end manner and greatly boost the performance of image SR. Afterward, many deep learning based methods have been proposed to improve the performance mainly by developing the network architecture. VDSR~\cite{vdsr} and IRCNN~\cite{ircnn} increased the network depth by adding more convolutional layers, and DRCN\cite{drcn} introduced recursive learning for parameter sharing. Tai~\emph{et al.} introduced recursive blocks in DRRN\cite{drrn} and memory block in Memnet\cite{memNet}. While all of these methods have greatly improved the SR performance by exploiting different network architecture, they have not considered the efficiency of SR, which lead to the learning based SR methods been away from application in reality. 

In contrast to chasing a smaller mean square error, we focus on the improvement of image restoration quality as well as boost the speed of the algorithm, which has been neglected for a long time. FSRCNN~\cite{fsrcnn} make an attempt to address this issue by adopting down-sampled patches as input and deconvolution to speed up the computing process. Their method effectively reduce redundancy and inspired us to explore the potential of accelerating SR. ESPCN\cite{espcn} used pixel shuffling operation to reduce features volume and checkerboard effect, which also greatly accelerated the SR network. Although these methods obtain a small running time, they don't fully utilize the inherent property of SR problem. For image SR, it has internal difficulty diversity, that is an area of an image with high frequency tend to lose more information during compressing while an area with low frequency tends to lose less. However, aforementioned methods ignore this property and tend to adopt a feed-forward model to process all samples.

\textbf{Neural network acceleration.} 
Obtaining a better balance between accuracy and efficiency has attracted many research communities for decades. Many studies have been proposed to change the connectivity structure of the deep convolutional networks such as ShuffleNet\cite{shufflenet} or introduce a more compact convolution operation such as in MobileNet\cite{mobilenet} and MobileNetV2\cite{mobilenetv2}. These studies have done great in reducing computation cost as well as maintain or even improve performance. However, these methods can be slower than a plain network in some computing platform. Some studies focus on reducing model size after training, such as weight pruning~\cite{lecun1990, li2016}, weight quantization \cite{hubara2016, rastegari2016}. These studies construct new models at the test time and re-train or fine-tune them to achieve a similar closer performance as the original models.

Other studies focus on alternating the evaluation manner. FractalNets\cite{fractalnet} perform prediction at any time by progressively evaluating subnetworks of the full network. Bolukbasi~\emph{et al.}~\cite{bolukbasi2017} addresses this problem by adaptively evaluating neural networks. Different from these works, MSDNet\cite{msdnet} adopts a specially designed network with multiple classifiers, which can directly output confidence scores to control the evaluation process for each test example. The adaptive computation time method \cite{graves2016} and its extension \cite{figurnov2017} also perform an adaptive evaluation of test examples but focus on skipping units rather than layers. Feedback Networks \cite{zamir2017} heavily shares parameters and allows early predictions in a recurrent process. However, their methods are less efficient in sharing computation. Our method is most inspired by MSDNet. Different from MSDNet, our proposed method focus on the difficulty diversity of image itself. And we also explore the frequency conflict occurred in a single model and therefore propose an original rolling strategy to handle the conflict.

\section{Methodology}

\begin{figure*}[htbp]
\centering

\label{fig:problem}
\subfloat[Successful Examples]{
\label{fig:successful}
\begin{minipage}[t]{0.5\textwidth}
\centering

\includegraphics[width=0.99\textwidth]{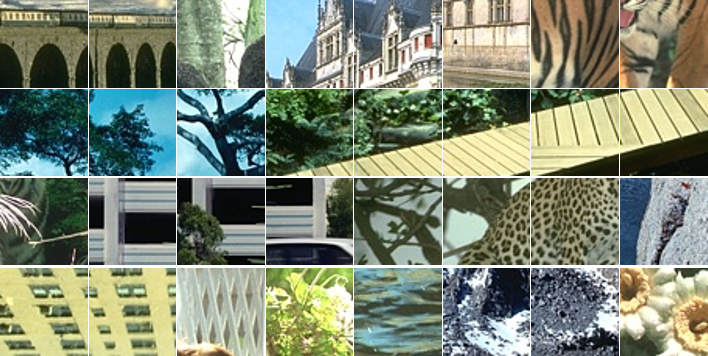}
\end{minipage}
}
\subfloat[Failure Examples]{
\label{fig:failure}
\begin{minipage}[t]{0.5\textwidth}
\centering

\includegraphics[width=0.99\textwidth]{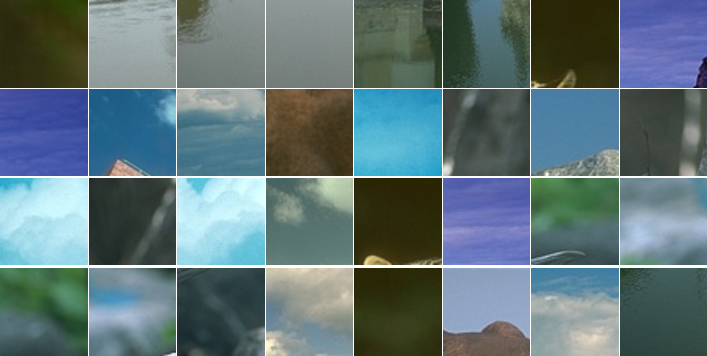}
\end{minipage}
}
\caption{Visualization results of a regular CNN for image SR. Left examples bring enormous PSNR(dB) promotion towards Bicubic. Right samples have higher absolute PSNR(dB) value as they contribute slight PSNR(dB) gain compared with Bicubic. It reveals that CNN is feasible to restore abundant texture images and bring great promotion, as patches with mild texture own a tiny upper bound.}
\end{figure*}

\begin{figure}[t]
\centering
\label{fig:feature}
\subfloat[]{
\label{fig:featurea}
\begin{minipage}[t]{0.48\textwidth}
\centering

\includegraphics[width=0.99\textwidth]{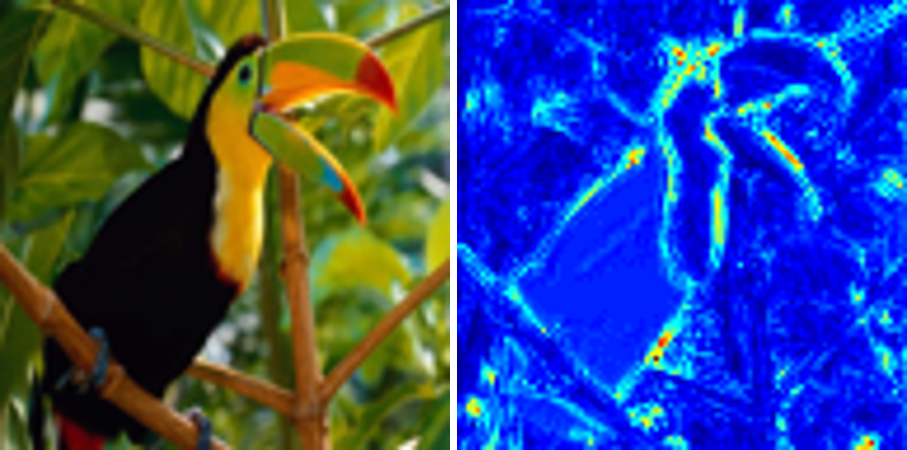}
\end{minipage}
}
\subfloat[]{
\label{fig:featureb}
\begin{minipage}[t]{0.48\textwidth}
\centering

\includegraphics[width=0.99\textwidth]{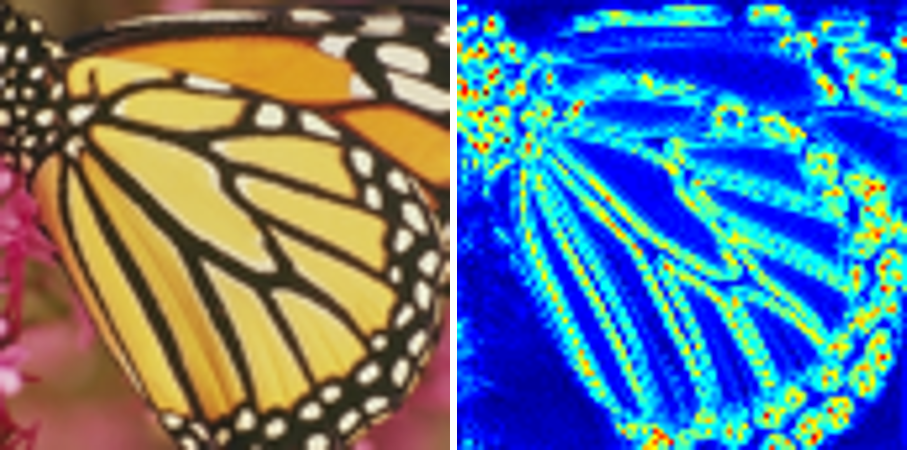}
\end{minipage}
}
\caption{The features of \emph{bird} and \emph{butterfly} are visualized in (a) and (b). For better visualization, we reduce dimension to 1 along channel with max operation.
}
\end{figure}

\textbf{Problem.} We investigate the failure cases which lead to poor performance in image SR. Given a 7-layers CNN, which has 6 convolutional layer with size of $3 \times 3 \times 32$ and a convolutional layer with size of $3 \times 3 \times 1$, we train it with 10 epochs on General-100~\cite{fsrcnn} to super-resolve images with factor $\times$3, we test it on BSDS100~\cite{bsd}. In Figure.\ref{fig:failure} and \ref{fig:successful}, we visualize its successful and failure examples. Meanwhile, we define the examples, which achieve more than 1dB improvement over Bicubic, as successful cases. The failure cases are the smaller images that obtain improvement less than 1dB. In Figure.\ref{fig:failure}, we can observe that the examples, which achieve minor improvement, are mild or inherently blurry. In contrast, the successful cases have a rich texture and drastic gradient. Moreover, we also extract feature in the middle of VDSR~\cite{vdsr} and present it in Fig.~\ref{fig:featurea} and~\ref{fig:featureb}. It can be observed that responses around high-frequency places are strong and VDSR gives low response toward the mild place.

According to our observation, we make three assumptions as follow. 1) The examples with rich texture can bring enormous gain. However, the mild examples are unable to demonstrate similar improvement. 2) Since deep neural network gives low response toward mild places and mild examples is intensely simple. A very deep neural network, which is widely used in image SR, owns a slight contribution to mild samples for further promotion. 3) The examples with severe, moderate and mild texture can be easily distinguished with its gradient information.  To address the aforementioned problems, we propose an end-to-end framework that joint learning image SR task with gradient prior knowledge.

\textbf{Overview of PRN.} The proposed PRN aims at learning a framework, which can super-resolve images more efficiently. More specific, the proposed framework first label patches according to gradient prior. Thus, we can fetch the different patches from different feature level. Since the bottom convolutional stage has tiny receptive field and the mild patches is different from severe samples in term of frequency, we then relieve these problems by adopting a novel strategy to roll convolutional filters. Next, we first describe the definition of gradient prior and then present the setting of the proposed framework.

\begin{figure}[t]
\centering
\label{fig:dis_problem}
\subfloat[]{
\label{fig:dis_successful}
\begin{minipage}[t]{0.49\textwidth}
\centering
\includegraphics[width=0.99\textwidth]{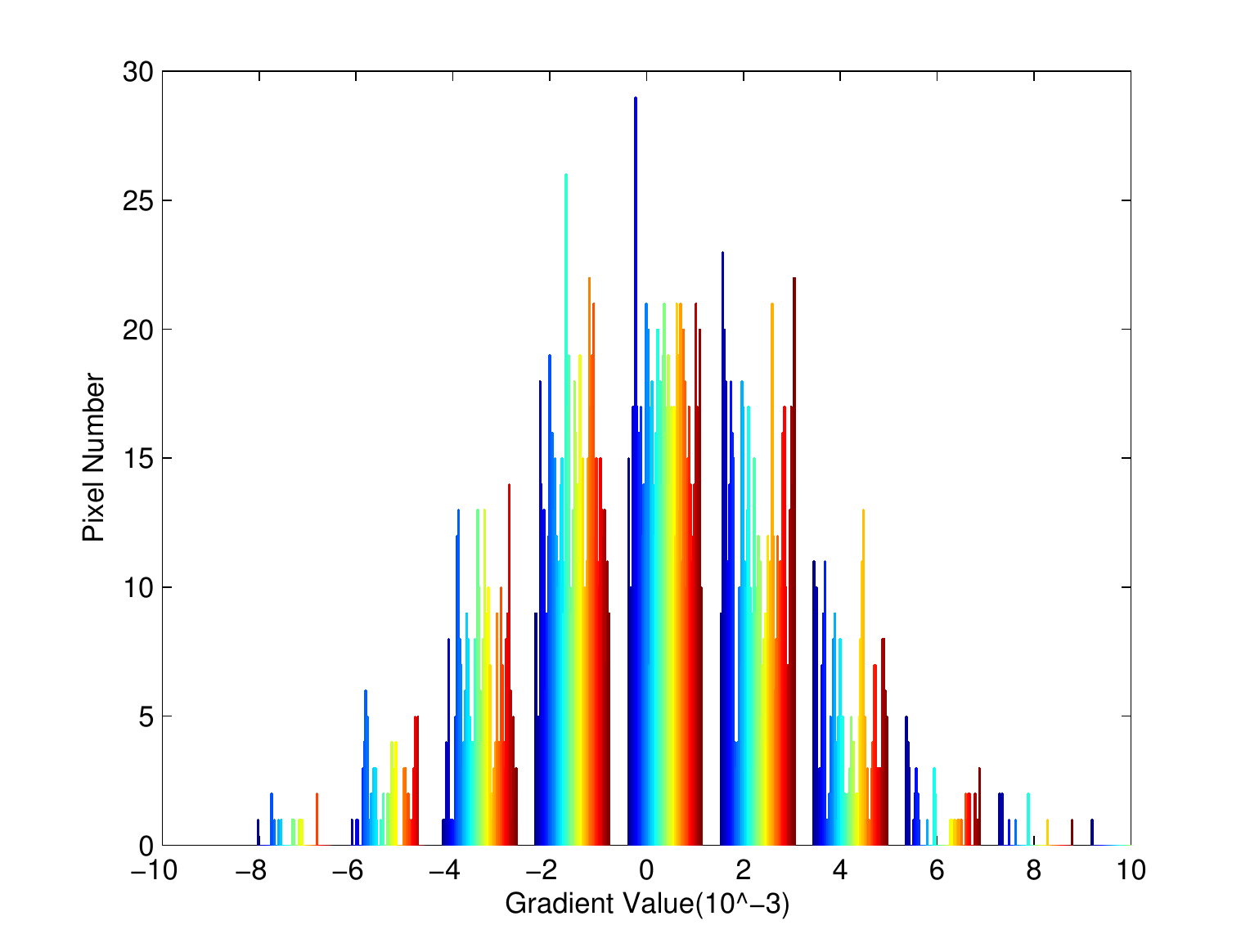}
\end{minipage}
}
\subfloat[]{
\label{fig:dis_failure}
\begin{minipage}[t]{0.49\textwidth}
\centering

\includegraphics[width=0.99\textwidth]{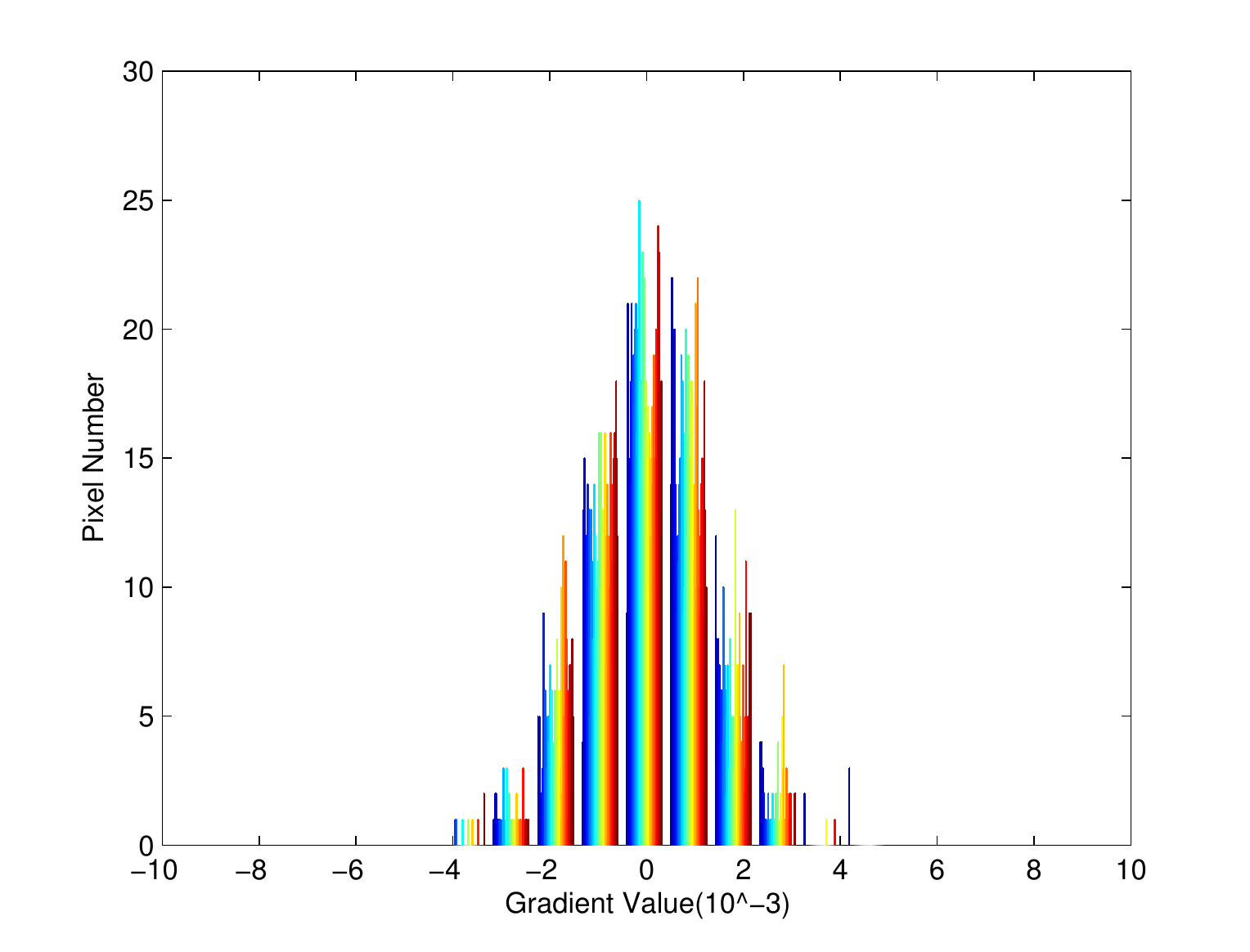}
\end{minipage}
}
\caption{Gradient properties of 10,000 severe and mild images. (a) Distribution of vertical axis gradient from high PSNR(dB) gain samples(e.g., severe images) in Figure.\ref{fig:successful}. (b) Distribution of vertical axis gradient from small PSNR(dB) gain cases(e.g., mild images) in Figure.~\ref{fig:failure}. We can easily distinguish successful and failure samples according to the gradient value.}
\end{figure}

\begin{figure*}[ht]
\centering
\includegraphics[width=0.98\textwidth]{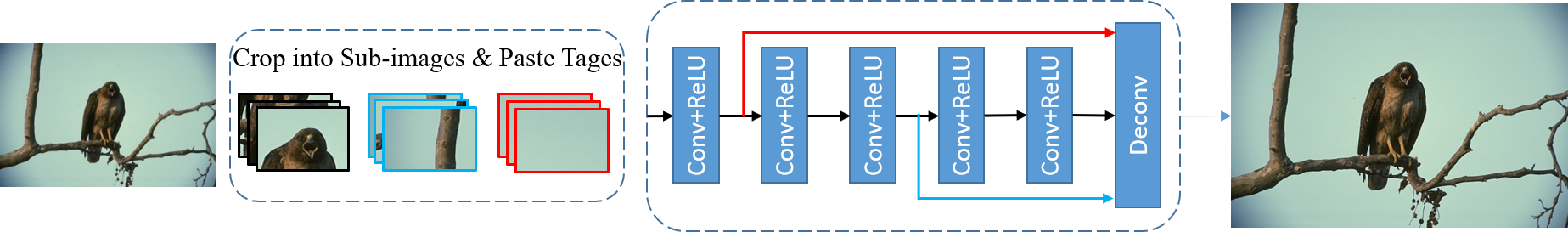}
\caption{Overview of the proposed PRN. The input image has plenty of blank regions and therefore we fetch results from the different position within the network to boost efficiency. We use \textbf{{\color{red} red}}, \textbf{{\color{blue} blue}}  and \textbf{black} bbox to label mild, moderate and severe patches and then send them into the network. The sub-images with different gradient prior will be fetched from the different convolutional layer. The arrow lines with the corresponding color indicate the stage where we fetch different patches. With this strategy, our method demonstrates a significant efficiency improvement.
}
\label{fig:framework}
\end{figure*}

\subsection{Gradient Prior} The proposed gradient prior is based on the observations that the failure samples in image SR usually have uniform gradient without sharp edges. As the samples with a uniform gradient contain rare pattern information and the upper bound for restoration is also pretty low, a simple and fast convolutional neural network can handle them well. We show the vertical gradient distribution of 10,000 successful and failure samples in Figure.\ref{fig:dis_successful} and \ref{fig:dis_failure}, respectively. It is obvious that mild samples have denser distribution among lower vertical gradient. And the distribution of severe images mainly lies on large value. With this gradient property, severe and mild samples can be distinguished. For an image, we describe the gradient property as follow:

\begin{equation}
\label{gradient_prior}
P(x)=\left \| G_{ver}(x) \right \|,
\end{equation}
where $x$ is the input image, $G_{ver}$ counts the gradient along the vertical axis. $P$ are the gradient prior knowledge, which also serves as a tag in our model. With the $P$, PRN is able to separate a set of images into mild, moderate and severe patches.

\begin{equation}
\begin{cases}
Mild                &  P \leq \gamma_{upper} \\
Moderate  & \gamma_{upper} \leq P \leq \gamma_{low} \\
Severe & \gamma_{low} \leq P.
\end{cases}
\end{equation}
The $\gamma_{upper}$ and $\gamma_{low}$ means the upper and low gradient threshold of $P$ for separating the images. Moreover, we make an ablation study on the gradient threshold in section~\ref{sec:ablation}. Although $P(x)$ is proposed based on the assumption that mild texture image is too simple to bring enormous gain, we show this prior can also be applied to accelerate image SR.

\subsection{Network architecture} As illustrated in Fig.\ref{fig:framework}, we put the patches with tag into the network for enhancement. To enable the network with a spacious receptive field, we use 64 convolutional kernels with a size of 5 $\times$ 5. To make full use of cuDNN~\cite{cudnn}, we employ 4 convolutional layers with 3 $\times$ 3 kernels and 64 channels. Before deconvolution operation, we conduct a shrinking layer with 64 kernels of size 1 $\times$ 1 to reduce parameters. Meanwhile, we add Leaky ReLU~\cite{leakyrelu} as activation function after each convolutional layer. Due to the efficient 3 $\times$ 3 kernels work on small size feature map directly, the proposed model can significantly accelerate the speed. At last, we use a deconvolution layer, whose stride is same to down-sampling factor, to perform an up-sampling operation. 

To obtain higher efficiency, we join auxiliary tag into our model by an end-to-end manner. In other words, the patch is able to be fetched from different feature level w.r.t tag knowledge. Since the mild patches are smooth and have less edge and texture, we prefer to obtain its feature from the first convolutional layer. Then, we use a deconvolution layer to obtain the restored patches. For moderate samples, we fetch from the third convolutional layer as they have a few texture and edge. A similar up-sampling operation also acts on the moderate samples for interpolation. Due to severe patches have rich texture and details, we conduct deconvolution layer on them after they forward all convolutional layers. The parameter of the network is optimized by $L_2$ loss. Meanwhile, different level of parameters is learned with different training pairs. For instance, the early stage layer is not only training with mild image pairs but also optimized with severe pairs. In contrast, the high-level parameter is optimized with severe pairs only. As shown in Fig.\ref{fig:framework}, we adopt such an efficient strategy to perform image SR.
\begin{figure*}[ht]
\centering
\includegraphics[width=0.98\textwidth]{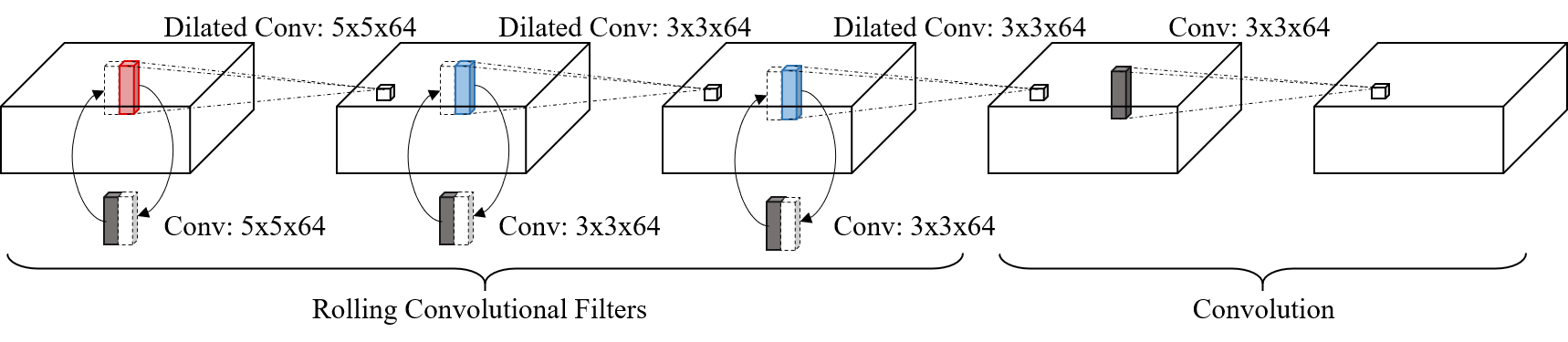}
\caption{Demonstration of the rolling strategy. The parameters for different patches(e.g., mild, moderate and severe) are label with \textbf{{\color{red} red}}, \textbf{{\color{blue} blue}}  and \textbf{black}. When we input the severe patch, the model roll the first three dilated convolutional layers(e.g., \textbf{{\color{red} red}} and \textbf{{\color{blue} blue}} columns)  with regular convolutional layers(e.g., \textbf{black} column). Nevertheless, suppose we send mild patch into the model, the model will replace dilated convolution(e.g., \textbf{{\color{red} red}}) with regular convolution layer(e.g., \textbf{black} column). With this content-adaptive and flexible strategy, our model improve the effectiveness and efficiency significantly.
}
\label{fig:rolling}
\end{figure*}
\subsection{Rolling the convolutional filters} 

Although we can effectively enhance the image with the aforementioned model, we still find the following problems: 1) The receptive field of the early stage is tiny. When we try to improve the performance, the tiny receptive field of early stage become a bottleneck. 2) Frequency conflicts. The frequency domain of mild and moderate examples are significantly different from severe patches. When we train mild samples in the early stage of the network, the output of high level is influenced. Thus, we make an attempt to resolve the above questions by developing a novel rolling strategy. 

Let $\Theta$ be parameters of a CNN, it consists of four parts of parameters $\left \{ \Theta_l,\Theta_m,\Theta_s,\Theta_{up} \right \}$. Meanwhile, $\Theta_l$ represents the early stage and consists of a convolutional layer. $\Theta_m$ means the middle stage and has two layers. $\Theta_s$ indicates the last stage and contains two convolutional layers and $\Theta_{up}$ is the parameters of a deconvolutional layer. In addition, we define auxiliary two set of the parameter $\Theta_l^D$ and $\Theta_m^D$. More specific, $\Theta_l^D$ represent a dilated convolution layer~\cite{dilation} with size of 64 $\times$ 5 $\times$ 5 and 1 dilation and $\Theta_m^D$ means two dilated convolution layers with size of 64 $\times$ 3 $\times$ 3 and 1 dilation as well. We use $\mathbb{P}_{HR}$ and $\mathbb{P}_{LR}$ represent high-res and low-res patch, respectively. Three superscripts $^{l}$, $^{m}$ and $^{s}$ are utilized to distinguish the mild, moderate and severe samples respectively. For instance, $\mathbb{P}_{LR}^{l}$ and $\mathbb{P}_{LR}^{s}$ indicate the low-res patch annotated with mild and severe tag. In sum, the enhancement toward severe patch can be defined as:
\begin{equation}
\mathbb{P}_{out} = f(\mathbb{P}_{LR}^{s} | {\Theta_l, \Theta _m, \Theta_s, \Theta_{up}}),
\end{equation}
where $f$ means the enhancement operation. As sketched in Fig.~\ref{fig:rolling}, the network will roll $\Theta_l$ and $\Theta_m$ with $\Theta_l^{D}$ and $\Theta_m^{D}$ and fetch the patch from different stage according to tag. More specific, suppose we input $\mathbb{P}_{LR}^{l}$ into the network, the model will enhance the patch with $\Theta_l^{D}$ explicit. By that analogy, the enhancement process toward $\mathbb{P}_{LR}^{l}$ and $\mathbb{P}_{LR}^{m}$ can be formulated as
\begin{equation}
\mathbb{P}_{out} = f(\mathbb{P}_{LR}^{l} | \Theta_l^D, \Theta_{up})
\end{equation}
and
\begin{equation}
\mathbb{P}_{out} = f(\mathbb{P}_{LR}^{m} | \Theta_l^D, \Theta_m^D, \Theta_{up}).
\end{equation}
With such flexible and content-adaptive rolling strategy, we not only resolve frequency conflicts but also increase the receptive field of early stage.

\subsection{End-to-end framework} In contrast to training models with different datasets, the proposed model not only be able to fetch images from the different stage but also optimize each stage with specific prior. The whole procedure can be formulated as an end-to-end framework to accelerate speed. We have sketch detailed algorithm in Algorithm.\ref{alg}. Since the down-scaled mild sample is similar to its ground-truth, the model is unable to learn how to recover realistic details and textures w.r.t mild training examples. To resolve this question, we adopt the mild and moderate samples as the training pairs for $\Theta_l^D$ and $\Theta_m^D$. With such an efficient strategy, our model not only greatly improve the performance but also accelerate the training and testing speed.

\begin{algorithm}
\caption{Learning Algorithm of PRN}
\label{alg}
\begin{algorithmic}[1]
\REQUIRE Training LR images $I_{LR}$; HR images $I_{HR}$;
\STATE Crop high-res and low-res images into patches $\mathbb{P}_{LR}$ and $\mathbb{P}_{HR}$ and distinguish $\mathbb{P}_{LR}$ into $\mathbb{P}_{LR}^l$, $\mathbb{P}_{LR}^m$ and $\mathbb{P}_{LR}^s$ w.r.t gradient prior;
\WHILE{$t < T$}
\STATE $t \leftarrow t+1$;
\STATE Choose a set of LR and HR patches, send low-res patches into network;
\STATE Obtain $ f(\mathbb{P}_{LR}^{m} | \Theta_l^D, \Theta_m^D, \Theta_{up})$, $f(\mathbb{P}_{LR}^{l} | \Theta_l^D, \Theta_{up})$ $f(\mathbb{P}_{LR}^{s} | {\Theta_l, \Theta _m, \Theta _s}, \Theta_{up})$ via forward propagation;
\STATE Update $\left \{ \Theta_l^D,\Theta_{up} \right \}$ with $\mathbb{P}_{LR}^{l}$ and $\mathbb{P}_{HR}$ pairs;
\STATE Update $\left \{ \Theta_l^D,\Theta_m^D, \Theta_{up} \right \} $ with $\mathbb{P}_{LR}^{m}$ and $\mathbb{P}_{HR}$ pairs;
\STATE Update $\left \{ \Theta_l, \Theta_m, \Theta_s, \Theta_{up} \right \} $ with $\mathbb{P}_{LR}^{s}$ and $\mathbb{P}_{HR}$ pairs;
\ENDWHILE  

\end{algorithmic}
\end{algorithm}

\section{Experiments}
\textbf{Datasets.} To make full use of the parameters in PRN, we use VOC2012~\cite{voc2012} to pre-train our model. VOC2012~\cite{voc2012} contains 17,125 clear images, which are taken from natural scene. Then, we finetune our model with BSD200~\cite{bsd}, which contains 200 images and is close to the real-world scene. BSD200~\cite{bsd} is augmented with scaling and rotation. We employ Set5, Set14, BSDS100, and Urban100 to evaluate our model.

\textbf{Implementation Details.} We use Xavier~\cite{xavier} initialize the parameters of the proposed model. Besides, the deconvolution layer is initialized according to the weight of Bicubic interpolation. We add pad with zero in each convolutional layer to assure the input tensor shares same size with the output. We convert all images from RGB to YCbCr and extract the Y channel for training. The training and testing images are cropped into 54$\times$54 patches and down-scaled with the corresponding factor to obtain the input. For 54 $\times$ 54 patch, the $\gamma_{upper}$ and $\gamma_{low}$ are set as $1\times 10^1$ and $3 \times 10^1$, respectively. In training, we set the batch size as 64 and learning rate is $1 \times 10^{-4}$ for all layers. In testing, we set the batch size as 1. The learning rate is reduced with factor $\times$10 for every 300 epochs. We use leaky ReLU with a negative slope of 0.2 as the activate function. We perform our training and testing on a desktop computer with i7-4790 CPU, GTX980Ti GPU, and 32GB RAM.

\textbf{Multi-scale training.} Different from some state-of-the-arts~\cite{fsrcnn,subpixel,srcnn}, which conduct its model with single factor training, we adopt multi-scale learning strategy to train PRN. Specifically, multi-scaling learning is to train the model with multiple down-sampling factors simultaneously. With the multi-scale learning, PRN can learn more contextual knowledge across different degeneration and achieves better performance.

\begin{table*}[]
\centering
\footnotesize
\caption{The PSNR and SSIM results of different approaches on Set5, Set14, BSDS100 and Urban100 with down-sampling factor $\times$2, $\times$3 and $\times$4. We use the \textbf{black} to label the firs place.}
\begin{tabular}{cccccccccc}
\hline

\multirow{2}{*}{Algorithm} & \multirow{2}{*}{Scale} & \multicolumn{2}{c}{Set5}                            & \multicolumn{2}{c}{Set14}                           & \multicolumn{2}{c}{BSDS100}                         & \multicolumn{2}{c}{URBAN100}                           \\
                           &                        & \multicolumn{1}{l}{PSNR} & \multicolumn{1}{l}{SSIM} & \multicolumn{1}{l}{PSNR} & \multicolumn{1}{l}{SSIM} & \multicolumn{1}{l}{PSNR} & \multicolumn{1}{l}{SSIM} & \multicolumn{1}{l}{PSNR} & \multicolumn{1}{l}{SSIM}  \\ \hline \hline
Bicubic                    & \multirow{8}{*}{2x}    & 33.69                    & 0.931                    & 30.25                    & 0.870                    & 29.57                    & 0.844                    & 26.89                    & 0.841                     \\
A+                         &                        & 36.60                    & 0.955                    & 32.32                    & 0.906                    & 31.24                    & 0.887                    & 29.25                    & 0.895                    \\
RFL                        &                        & 36.59                    & 0.954                    & 32.29                    & 0.905                    & 31.18                    & 0.885                    & 29.14                    & 0.891                     \\
SelfEx                     &                        & 36.60                    & 0.955                    & 32.24                    & 0.904                    & 31.20                    & 0.887                    & 29.55                    & 0.898                     \\
SRCNN                      &                        & 36.72                    & 0.955                    & 32.51                    & 0.908                    & 31.38                    & 0.889                    & 29.53                    & 0.896                    \\
SCN                        &                        & 36.58                    & 0.954                    & 32.35                    & 0.905                    & 31.26                    & 0.885                    & 29.52                    & 0.897                    \\
FSRCNN                     &                        & 37.05                    & 0.956                    & 32.66                    & 0.909                    & 31.53                    & 0.892                    & 29.88                    & 0.902                     \\
Our                        &                        & \textbf{37.09}                     &\textbf{0.957}                    & \textbf{32.90}                    &\textbf{0.910}                     &\textbf{31.66}                     &\textbf{0.893}                    &\textbf{30.23}                  &\textbf{0.909}                     \\ \hline \hline
Bicubic                    & \multirow{8}{*}{3x}    & 30.41                    & 0.869                    & 27.55                    & 0.775                    & 27.22                    & 0.741                    & 24.47                    & 0.737                     \\
A+                         &                        & 32.62                    & 0.909                    & 29.15                    & 0.820                    & 28.31                    & 0.785                    & 26.05                    & 0.799                    \\
RFL                        &                        & 32.47                    & 0.906                    & 29.07                    & 0.818                    & 28.23                    & 0.782                    & 25.88                    & 0.792                     \\
SelfEx                     &                        & 32.66                    & 0.910                    & 29.18                    & 0.821                    & 28.30                    & 0.786                    & 26.45                    & 0.810                    \\
SRCNN                      &                        & 32.78                    & 0.909                    & 29.32                    & 0.823                    & 28.42                    & 0.788                    & 26.25                    & 0.801                    \\
SCN                        &                        & 32.62                    & 0.908                    & 29.16                    & 0.818                    & 28.33                    & 0.783                    & 26.21                    & 0.801                    \\
FSRCNN                     &                        & 33.18                    & 0.914                    & 29.37                    & 0.824                    & 28.53                    & 0.791                    & 26.43                    & 0.808                     \\
Our                        &                        &\textbf{33.32}                     &\textbf{0.916}                     &\textbf{29.64}                     &\textbf{0.828}                     &\textbf{28.72}                     &\textbf{0.794}                     &\textbf{26.75}                     &\textbf{0.815}                    \\ \hline \hline
Bicubic                    & \multirow{8}{*}{4x}    & 28.43                    & 0.811                    & 26.01                    & 0.704                    & 25.97                    & 0.670                    & 23.15                    & 0.660                     \\
A+                         &                        & 30.32                    & 0.860                    & 27.34                    & 0.751                    & 26.83                    & 0.711                    & 24.34                    & 0.721                                        \\
RFL                        &                        & 30.17                    & 0.855                    & 27.24                    & 0.747                    & 26.76                    & 0.708                    & 24.20                    & 0.712                    \\
SelfEx                     &                        & 30.34                    & 0.862                    & 27.41                    & 0.753                    & 26.84                    & 0.713                    & 24.83                    & 0.740                                        \\
SRCNN                      &                        & 30.50 
& 0.863                    &27.52                     &0.753                     &26.91                     &0.712                     &24.53                     &0.725                     \\
SCN                        &                        &30.41                     &0.863                     &27.39                     &0.751                     &26.88                     &0.711                     &24.52                     &0.726                      \\
FSRCNN                     &                        &30.72                     &0.866                     &27.61                     &0.755                     &26.98                     &0.715                     &24.62                     &0.728                     \\
Our                        &                        &\textbf{31.08}                     &\textbf{0.875}                     &\textbf{27.89}                     &\textbf{0.762}                     &\textbf{27.17}                     &\textbf{0.728}                     &\textbf{24.86}                     &\textbf{0.733}                      \\ \hline
\end{tabular}
\label{tab:comparison}
\end{table*}

\begin{table}
\centering
\footnotesize
\caption{We have compared a wide range of potential gradient threshold. Meanwhile, L indicates $\gamma_{low}$ and U is $\gamma_{upper}$. The suffix number along L and U means different threshold value.}
\begin{tabular}{lllllllll} 
\toprule
$\gamma$               & L1 & L2 & L3 & L4 & U1 & U2 & U3 & U4  \\ 
\cline{5-9}
\hline
Value(10$\times$1) & 1  & 2  & 5  & 7  & 3  & 5  & 8  & 10  \\
\bottomrule
\end{tabular}
\label{tab:depth}
\end{table}

\subsection{Comparison with State-of-the-arts.} We compare our model with state-of-the-art methods, including A+~\cite{aplus}, SRF~\cite{srf}, SelfEx~\cite{selfsr}, RFL~\cite{rfl}, SCN~\cite{scn}, SRCNN~\cite{srcnn}, LapSRN~\cite{laplacian}, VDSR~\cite{vdsr}, DRCN~\cite{drcn}, and FSRCNN~\cite{fsrcnn}. We adopt widely used quality metrics, e.g., PSNR and SSIM, to evaluate our model. For DRCN~\cite{drcn}, we use our own implementation for comparison. For rest of other methods, we use their public code and model to obtain results.

As shown in table.~\ref{tab:comparison}, our model achieve superior performance among light-weight methods~\cite{aplus,srf,selfsr,rfl,scn,srcnn,fsrcnn}. Compared with FSRCNN, our model achieve 0.13 dB and 0.19 dB promotion on BSDS100 with factor 2$\times$ and 3$\times$. Similarly, our model obtains 0.69 dB gain when compared with FSRCNN on Mange109 with factor 4$\times$. With the limitation of the parameter, our model is weak than heavy inferences~\cite{drcn,vdsr,laplacian}. As sketched in figure.~\ref{fig:efficiency}, our model shows slightly lower performance compared with huge model~\cite{drcn,vdsr,laplacian}, but our speed is accelerated about several times. Therefore, the model is particularly competitive for mobile devices and applications.

\begin{figure*}[]
\begin{minipage}{0.19\linewidth}
\centerline{\includegraphics[width=1\textwidth]{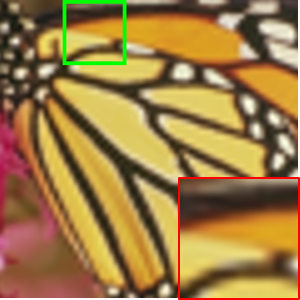}}
\centerline{Bicubic}
\centerline{22.18 dB}
\end{minipage}
\hfill 
\begin{minipage}{0.19\linewidth}
\centerline{\includegraphics[width=1\textwidth]{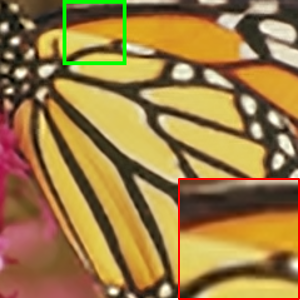}}
\centerline{A+~\cite{aplus}}
\centerline{24.65 dB}
\end{minipage}
\hfill
\begin{minipage}{0.19\linewidth}
\centerline{\includegraphics[width=1\textwidth]{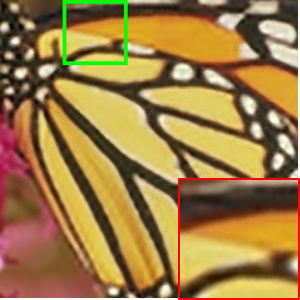}}
\centerline{RFL~\cite{rfl}}
\centerline{24.56 dB}
\end{minipage}
\hfill
\begin{minipage}{0.19\linewidth}
\centerline{\includegraphics[width=1\textwidth]{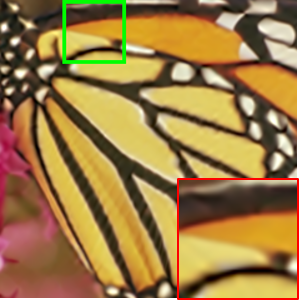}}
\centerline{SelfExSR~\cite{selfsr}}
\centerline{24.29 dB}
\end{minipage}
\hfill
\begin{minipage}{0.19\linewidth}
\centerline{\includegraphics[width=1\textwidth]{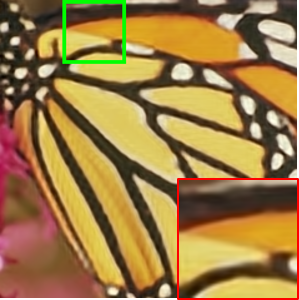}}
\centerline{SRCNN~\cite{srcnn}}
\centerline{25.65 dB}
\end{minipage}
\vfill
\vspace{3mm}
\begin{minipage}{0.19\linewidth}
\centerline{\includegraphics[width=1\textwidth]{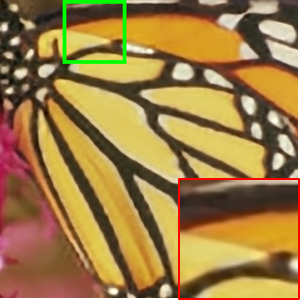}}
\centerline{SCN~\cite{scn}}
\centerline{25.51 dB}
\end{minipage}
\hfill
\begin{minipage}{0.19\linewidth}
\centerline{\includegraphics[width=1\textwidth]{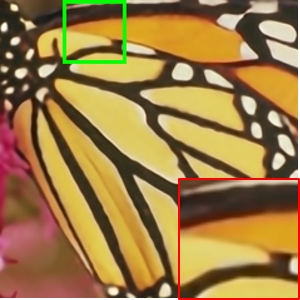}}
\centerline{{\color{red} {LapSRN}}~\cite{laplacian}}
\centerline{27.55 dB}
\end{minipage}
\hfill
\begin{minipage}{0.19\linewidth}
\centerline{\includegraphics[width=1\textwidth]{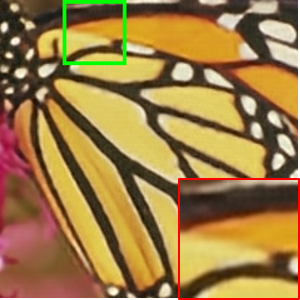}}
\centerline{FSRCNN~\cite{fsrcnn}}
\centerline{25.90 dB}
\end{minipage}
\hfill
\begin{minipage}{0.19\linewidth}
\centerline{\includegraphics[width=1\textwidth]{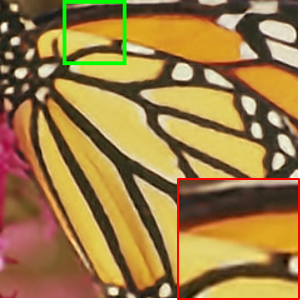}}
\centerline{{\color{blue} Our}}
\centerline{26.71 dB}
\end{minipage}
\hfill
\begin{minipage}{0.19\linewidth}
\centerline{\includegraphics[width=1\textwidth]{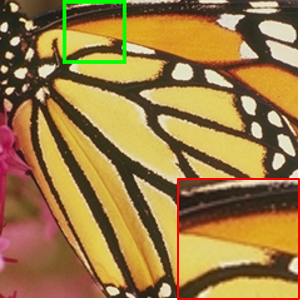}}
\centerline{Original}
\centerline{}
\end{minipage}
\caption{Qualitative comparison on 'butterfly' with the scaling factor of 4. We use {\color{red} red} and {\color{blue} blue} to label best two results, respectively. Best viewed by zooming in the electronic version.}
\label{fig:butterfly}
\end{figure*}

\begin{figure*}[]
\begin{minipage}{0.19\linewidth}
\centerline{\includegraphics[width=1\textwidth]{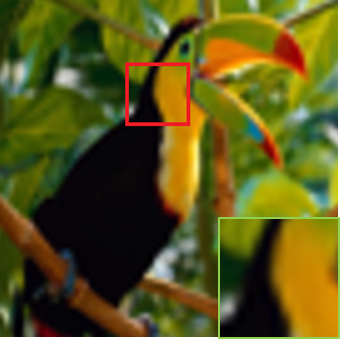}}
\centerline{Bicubic}
\centerline{30.22 dB}
\end{minipage}
\hfill 
\begin{minipage}{0.19\linewidth}
\centerline{\includegraphics[width=1\textwidth]{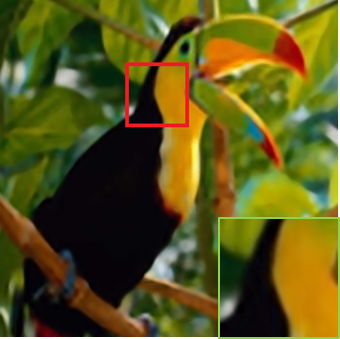}}
\centerline{A+~\cite{aplus}}
\centerline{32.63 dB}
\end{minipage}
\hfill
\begin{minipage}{0.19\linewidth}
\centerline{\includegraphics[width=1\textwidth]{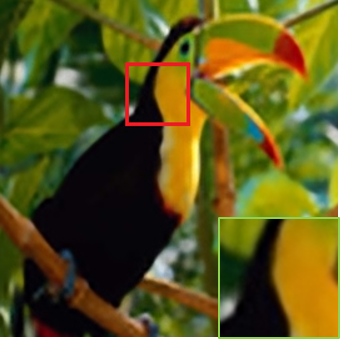}}
\centerline{RFL~\cite{rfl}}
\centerline{32.33 dB}
\end{minipage}
\hfill
\begin{minipage}{0.19\linewidth}
\centerline{\includegraphics[width=1\textwidth]{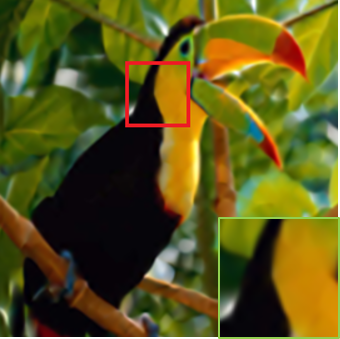}}
\centerline{SelfExSR~\cite{selfsr}}
\centerline{32.90 dB}
\end{minipage}
\hfill
\begin{minipage}{0.19\linewidth}
\centerline{\includegraphics[width=1\textwidth]{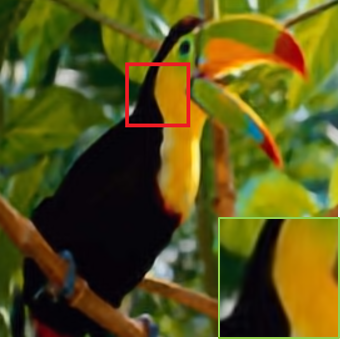}}
\centerline{SRCNN~\cite{srcnn}}
\centerline{32.61 dB}
\end{minipage}
\vfill
\vspace{3mm}
\begin{minipage}{0.19\linewidth}
\centerline{\includegraphics[width=1\textwidth]{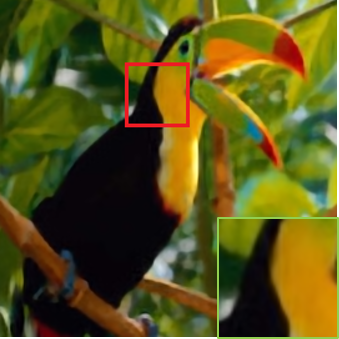}}
\centerline{SCN~\cite{scn}}
\centerline{32.47 dB}
\end{minipage}
\hfill
\begin{minipage}{0.19\linewidth}
\centerline{\includegraphics[width=1\textwidth]{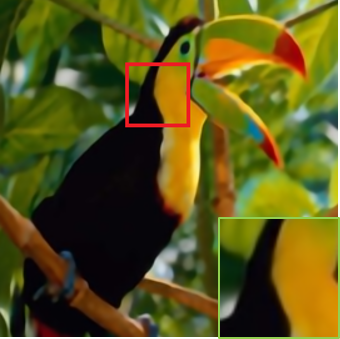}}
\centerline{{\color{red} {LapSRN}}~\cite{laplacian}}
\centerline{33.82 dB}
\end{minipage}
\hfill
\begin{minipage}{0.19\linewidth}
\centerline{\includegraphics[width=1\textwidth]{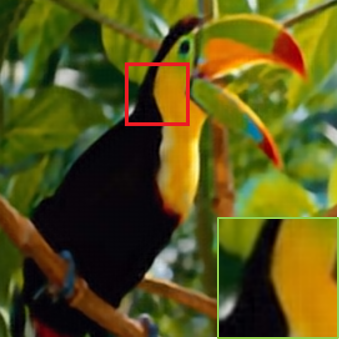}}
\centerline{FSRCNN~\cite{fsrcnn}}
\centerline{32.86 dB}
\end{minipage}
\hfill
\begin{minipage}{0.19\linewidth}
\centerline{\includegraphics[width=1\textwidth]{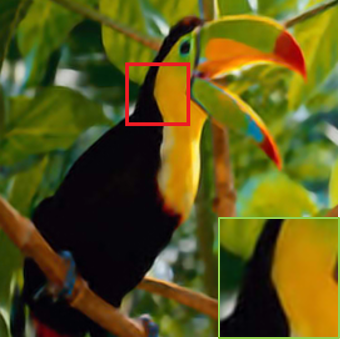}}
\centerline{{\color{blue} Our}}
\centerline{33.16 dB}
\end{minipage}
\hfill
\begin{minipage}{0.19\linewidth}
\centerline{\includegraphics[width=1\textwidth]{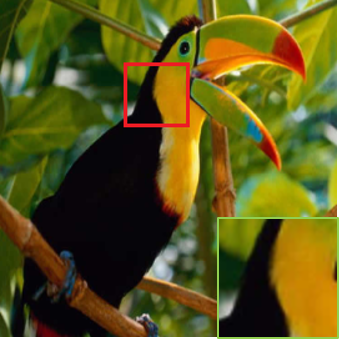}}
\centerline{Original}
\centerline{}
\end{minipage}
\caption{Qualitative comparison on 'bird' with the scaling factor of 4. We use {\color{red} red} and {\color{blue} blue} to label best two results, respectively. Best viewed by zooming in the electronic version.}
\label{fig:bird}
\end{figure*}

\begin{figure*}[]
\begin{minipage}{0.19\linewidth}
\centerline{\includegraphics[width=1\textwidth]{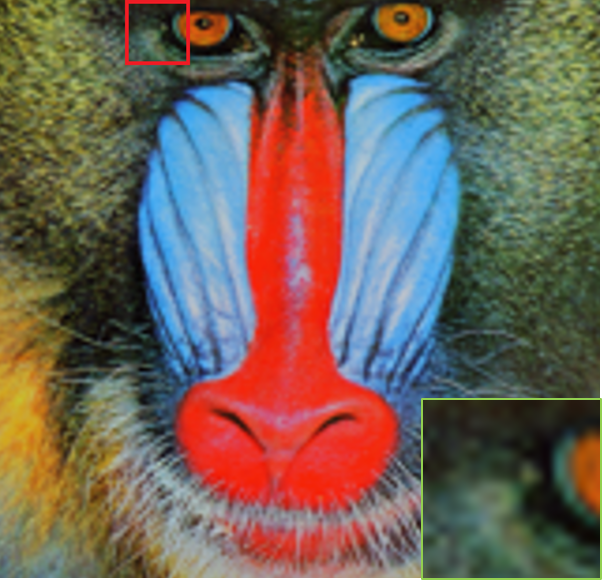}}
\centerline{Bicubic}
\centerline{23.19 dB}
\end{minipage}
\hfill 
\begin{minipage}{0.19\linewidth}
\centerline{\includegraphics[width=1\textwidth]{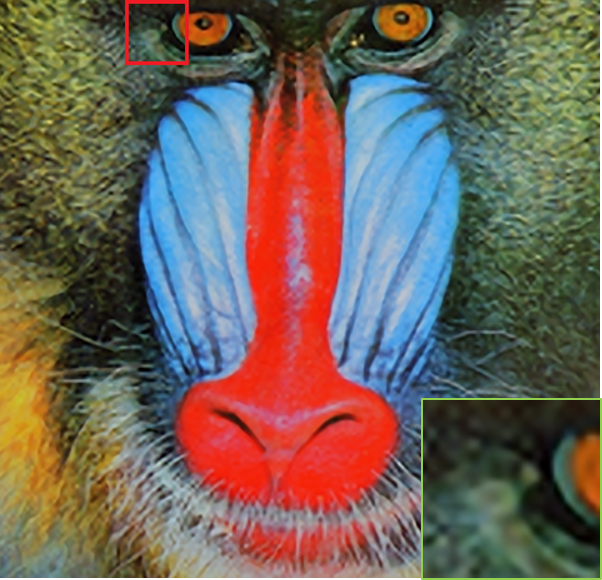}}
\centerline{A+~\cite{aplus}}
\centerline{23.62 dB}
\end{minipage}
\hfill
\begin{minipage}{0.19\linewidth}
\centerline{\includegraphics[width=1\textwidth]{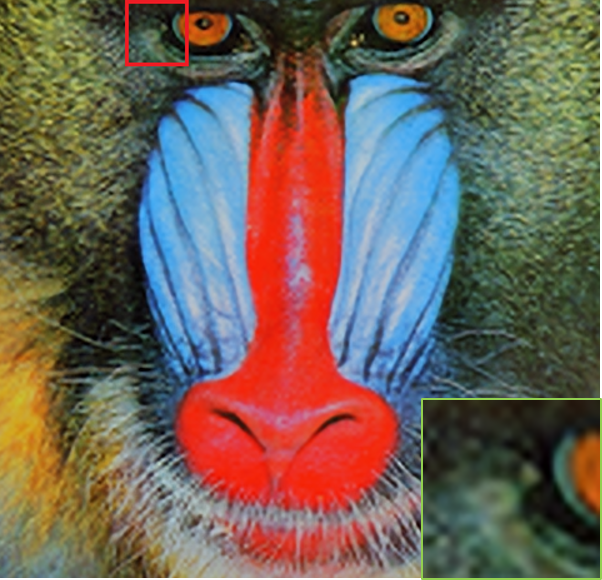}}
\centerline{RFL~\cite{rfl}}
\centerline{23.59 dB}
\end{minipage}
\hfill
\begin{minipage}{0.19\linewidth}
\centerline{\includegraphics[width=1\textwidth]{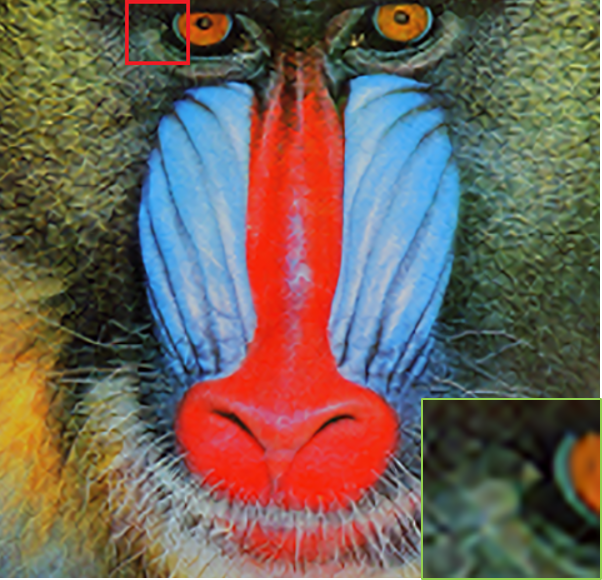}}
\centerline{SelfExSR~\cite{selfsr}}
\centerline{23.51 dB}
\end{minipage}
\hfill
\begin{minipage}{0.19\linewidth}
\centerline{\includegraphics[width=1\textwidth]{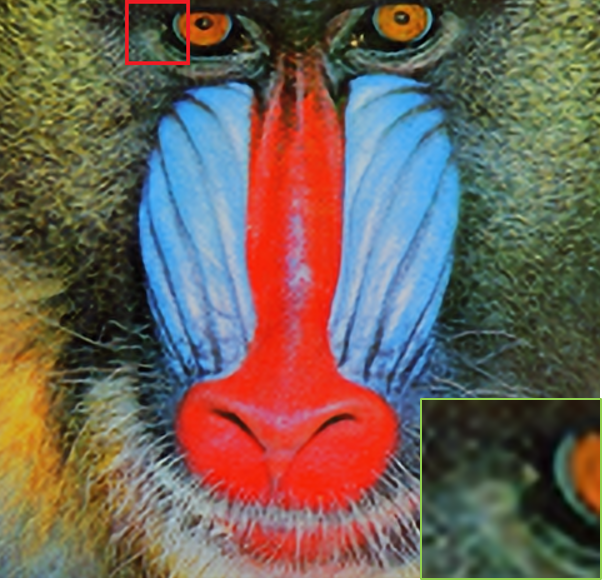}}
\centerline{SRCNN~\cite{srcnn}}
\centerline{23.67 dB}
\end{minipage}
\vfill
\vspace{3mm}
\begin{minipage}{0.19\linewidth}
\centerline{\includegraphics[width=1\textwidth]{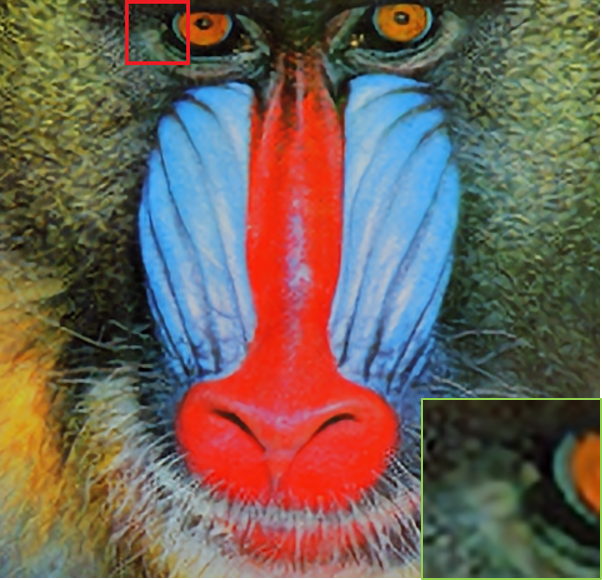}}
\centerline{SCN~\cite{scn}}
\centerline{23.60 dB}
\end{minipage}
\hfill
\begin{minipage}{0.19\linewidth}
\centerline{\includegraphics[width=1\textwidth]{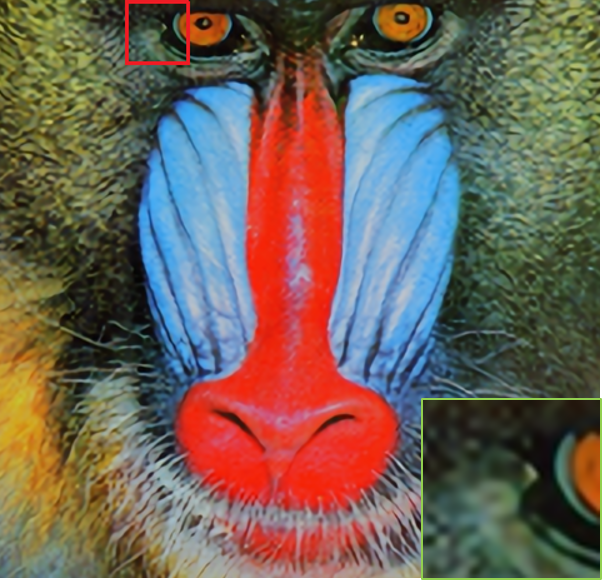}}
\centerline{{\color{blue} {LapSRN}}~\cite{laplacian}}
\centerline{23.74 dB}
\end{minipage}
\hfill
\begin{minipage}{0.19\linewidth}
\centerline{\includegraphics[width=1\textwidth]{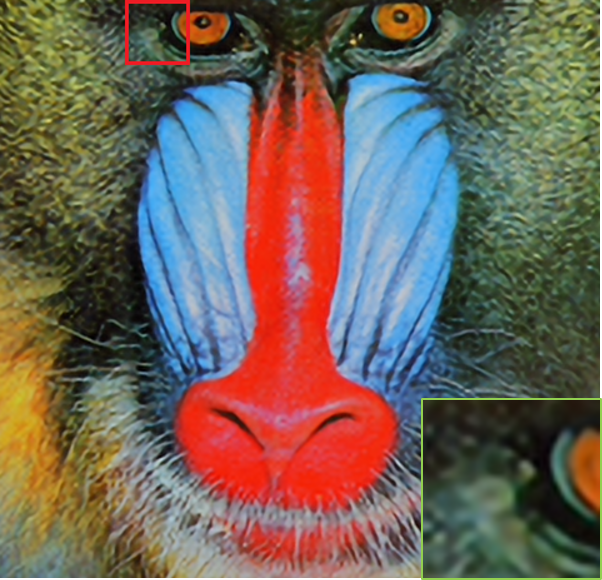}}
\centerline{FSRCNN~\cite{fsrcnn}}
\centerline{23.64 dB}
\end{minipage}
\hfill
\begin{minipage}{0.19\linewidth}
\centerline{\includegraphics[width=1\textwidth]{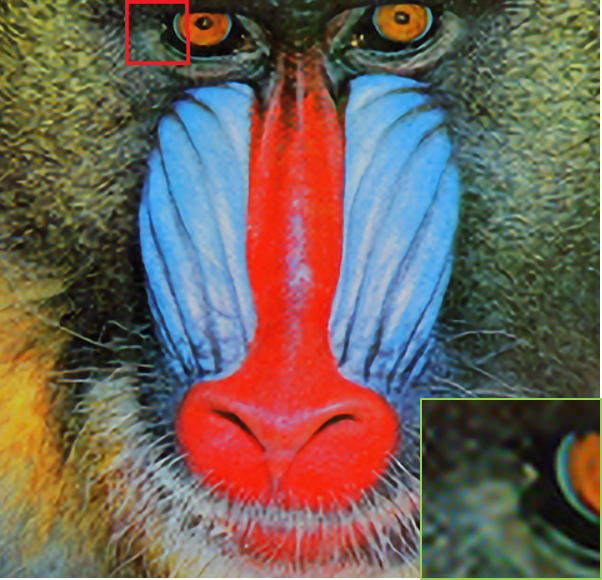}}
\centerline{{\color{red} Our}}
\centerline{23.75 dB}
\end{minipage}
\hfill
\begin{minipage}{0.19\linewidth}
\centerline{\includegraphics[width=1\textwidth]{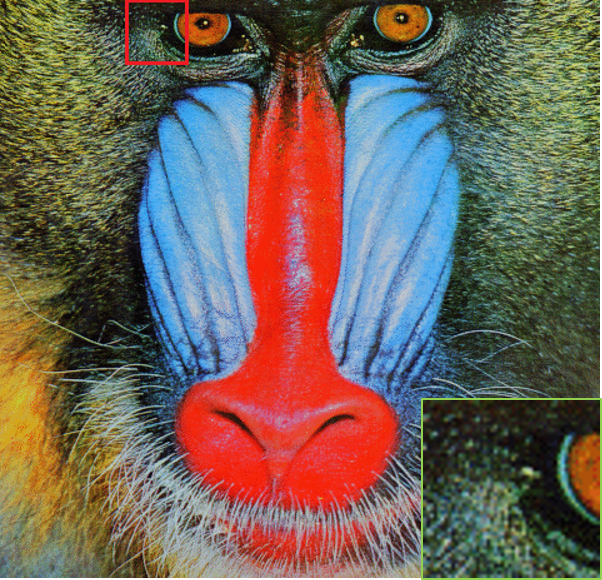}}
\centerline{Original}
\centerline{}
\end{minipage}
\caption{Qualitative comparison on 'baboon' with the scaling factor of 3. We use {\color{red} red} and {\color{blue} blue} to label best two results, respectively. Best viewed by zooming in the electronic version.}
\label{fig:baboon}
\end{figure*}

\begin{figure*}[]
\begin{minipage}{0.19\linewidth}
\centerline{\includegraphics[width=1\textwidth]{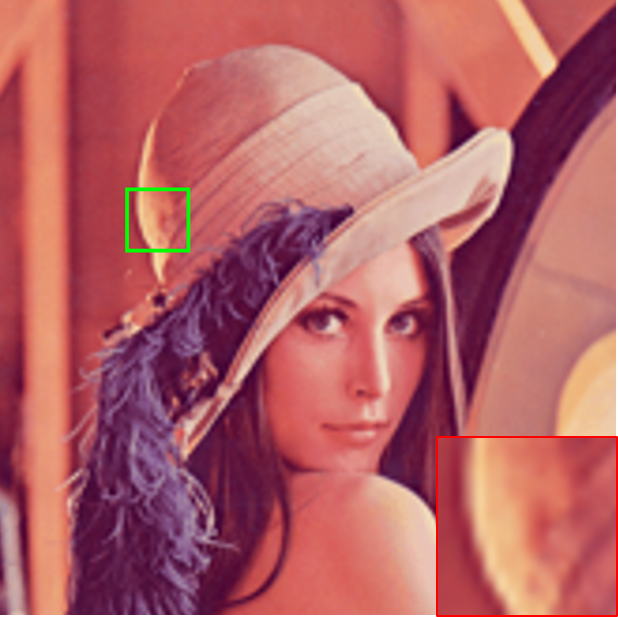}}
\centerline{Bicubic}
\centerline{31.54 dB}
\end{minipage}
\hfill 
\begin{minipage}{0.19\linewidth}
\centerline{\includegraphics[width=1\textwidth]{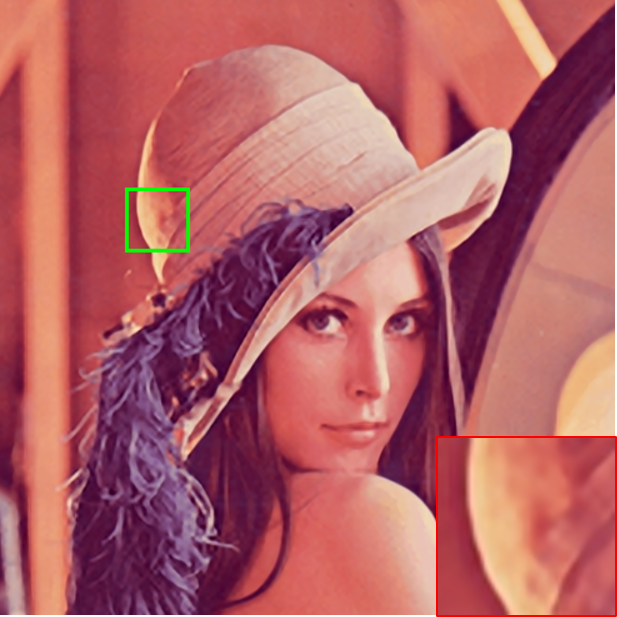}}
\centerline{A+~\cite{aplus}}
\centerline{33.41 dB}
\end{minipage}
\hfill
\begin{minipage}{0.19\linewidth}
\centerline{\includegraphics[width=1\textwidth]{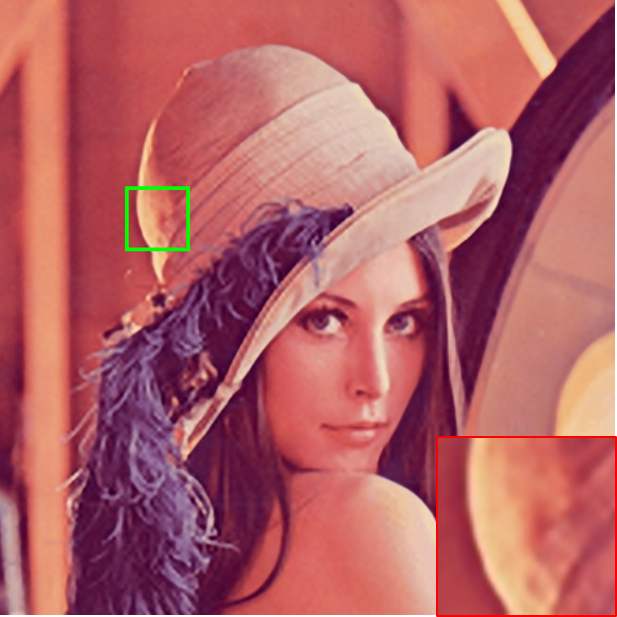}}
\centerline{RFL~\cite{rfl}}
\centerline{33.33 dB}
\end{minipage}
\hfill
\begin{minipage}{0.19\linewidth}
\centerline{\includegraphics[width=1\textwidth]{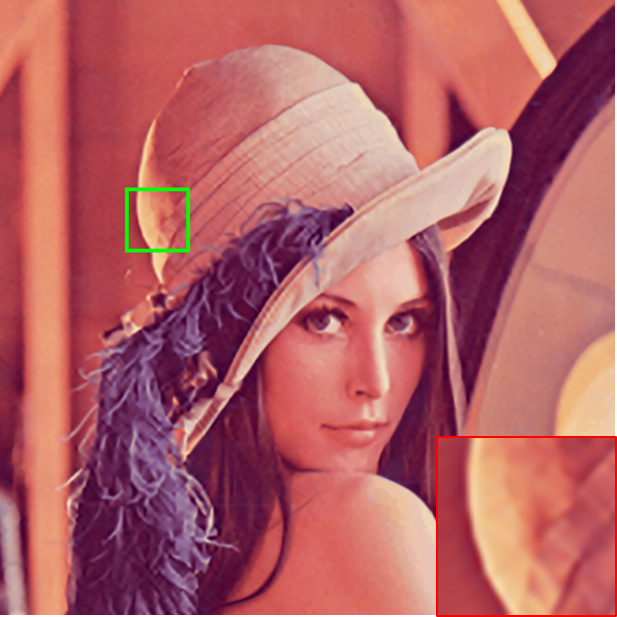}}
\centerline{SelfExSR~\cite{selfsr}}
\centerline{33.40 dB}
\end{minipage}
\hfill
\begin{minipage}{0.19\linewidth}
\centerline{\includegraphics[width=1\textwidth]{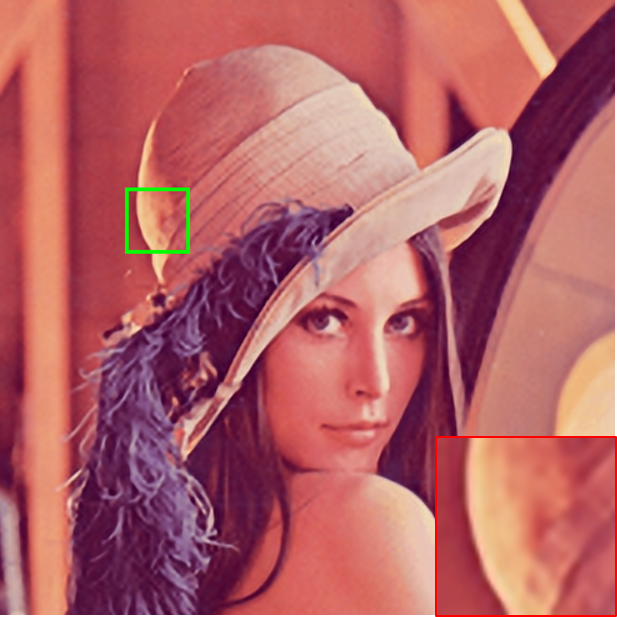}}
\centerline{SRCNN~\cite{srcnn}}
\centerline{33.55 dB}
\end{minipage}
\vfill
\vspace{3mm}
\begin{minipage}{0.19\linewidth}
\centerline{\includegraphics[width=1\textwidth]{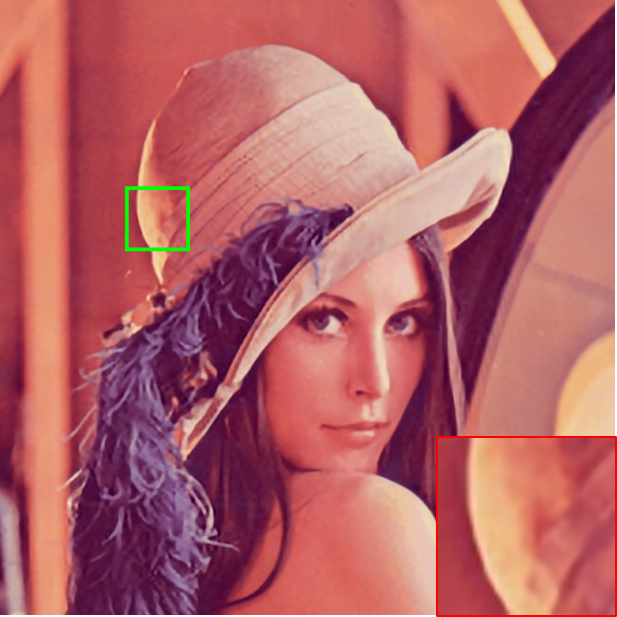}}
\centerline{SCN~\cite{scn}}
\centerline{33.36 dB}
\end{minipage}
\hfill
\begin{minipage}{0.19\linewidth}
\centerline{\includegraphics[width=1\textwidth]{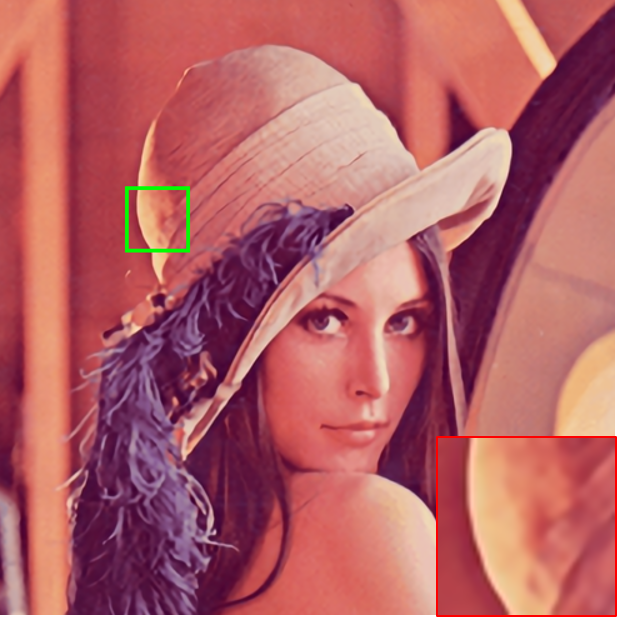}}
\centerline{{\color{red} {LapSRN}}~\cite{laplacian} }
\centerline{33.88 dB}
\end{minipage}
\hfill
\begin{minipage}{0.19\linewidth}
\centerline{\includegraphics[width=1\textwidth]{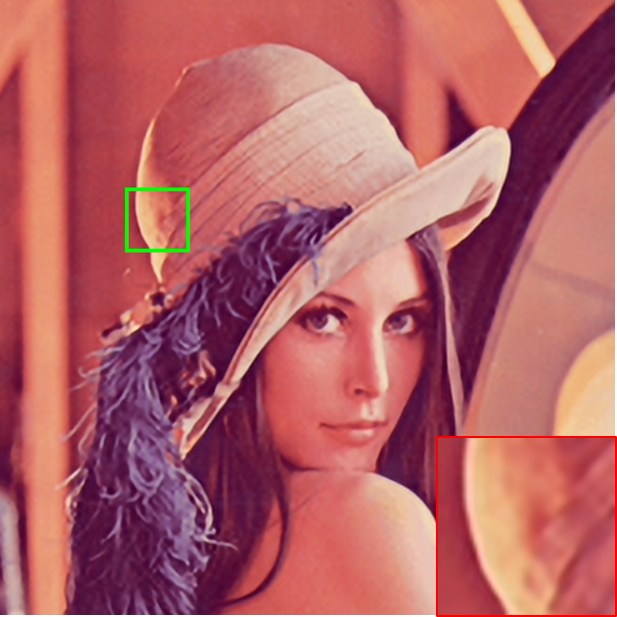}}
\centerline{FSRCNN~\cite{fsrcnn} }
\centerline{33.59 dB}
\end{minipage}
\hfill
\begin{minipage}{0.19\linewidth}
\centerline{\includegraphics[width=1\textwidth]{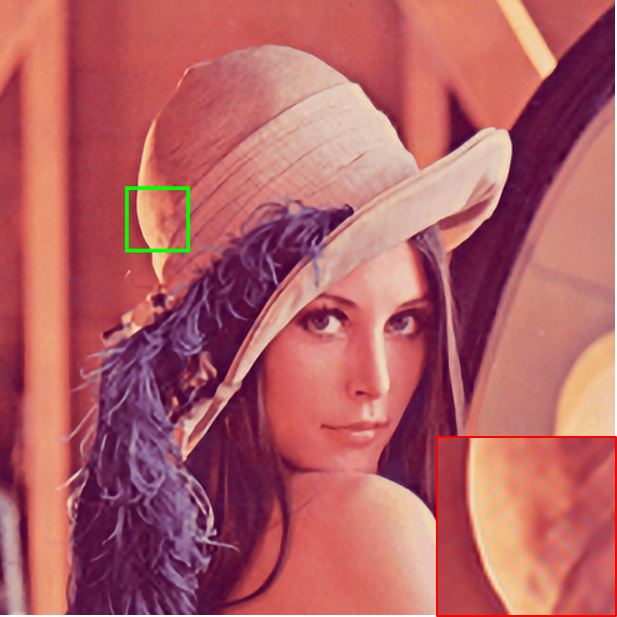}}
\centerline{{\color{blue} Our} }
\centerline{33.67 dB}
\end{minipage}
\hfill
\begin{minipage}{0.19\linewidth}
\centerline{\includegraphics[width=1\textwidth]{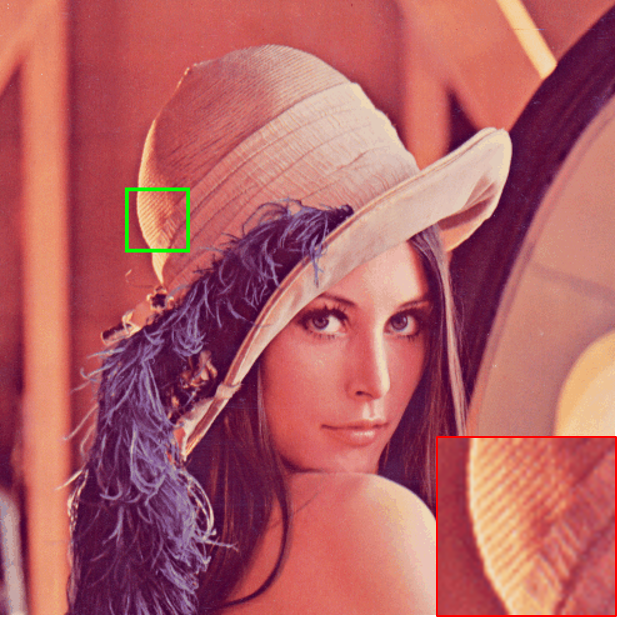}}
\centerline{Original}
\centerline{}
\end{minipage}
\caption{Qualitative comparison on 'lena' with the scaling factor of 3. We use {\color{red} red} and {\color{blue} blue} to label best two results, respectively. Best viewed by zooming in the electronic version.}
\label{fig:lena}
\end{figure*}

We also show qualitative comparison in Figure.~\ref{fig:butterfly},~\ref{fig:bird},~\ref{fig:baboon} and~\ref{fig:lena}. For better visualization, we interpolate the chrominance space by bicubic to obtain color images. Compared with other methods, our approach can generate image clearer boundary and rich details.

\subsection{Ablation study}
\label{sec:ablation}
In this section, we mainly investigate different settings of the proposed model and provide insights into the choice of hyper-parameters.
\begin{figure}[ht]
\centering
\includegraphics[width=0.88\textwidth]{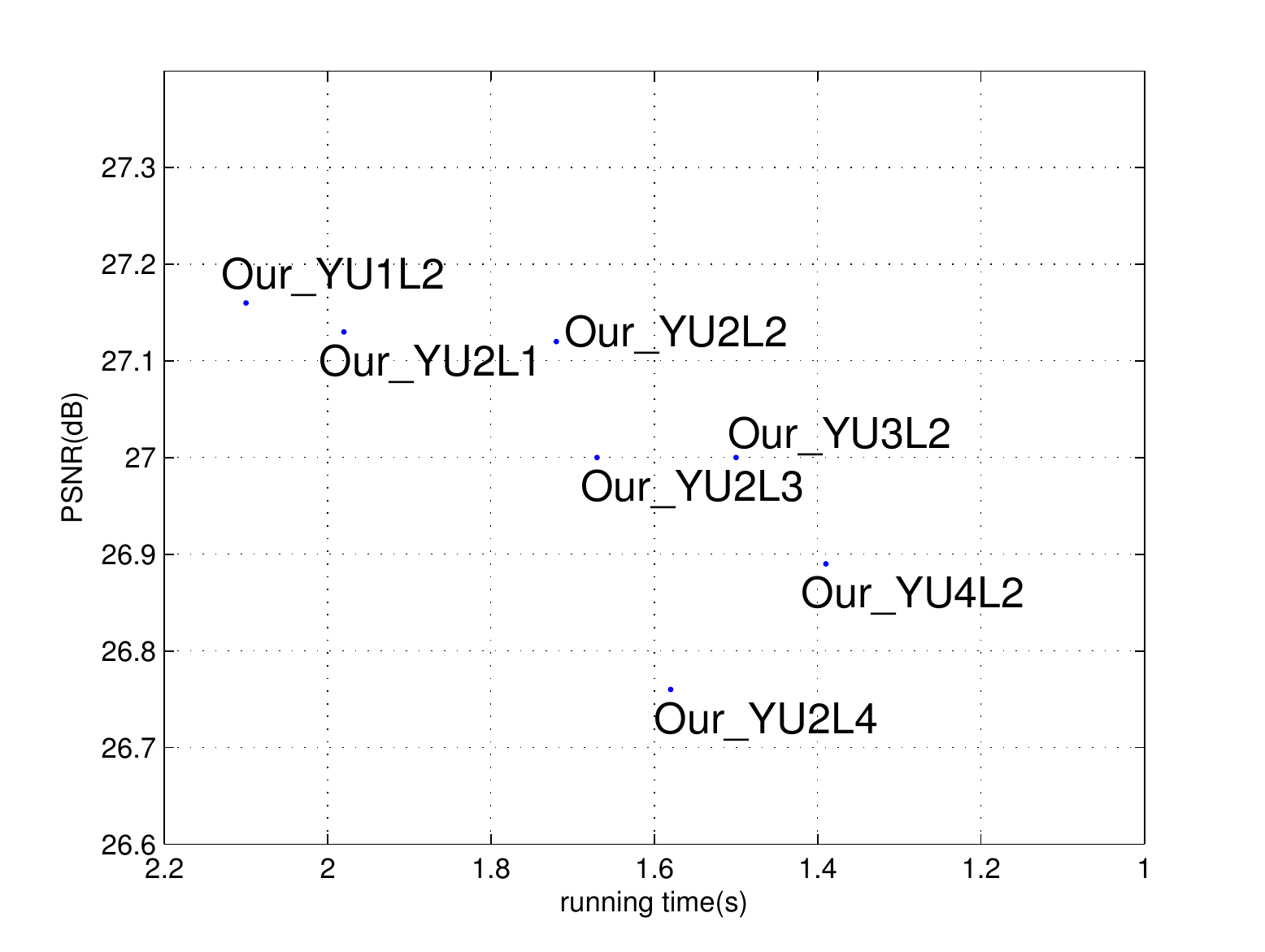}
\caption{Efficiency and effectiveness analysis of different gradient threshold on BSDS100. }
\label{fig:threshold}
\end{figure}

%%%%%%%%%%%%%%%%%Figure for butterfly%%%%%%%%%%%%%%%%

%%%%%%%%%%%%%%%%%Figure for bird%%%%%%%%%%%%%%%%

%%%%%%%%%%%%%%%%%Figure for baboon%%%%%%%%%%%%%%%%

%%%%%%%%%%%%%%%%%Figure for lena%%%%%%%%%%%%%%%%

\textbf{Gradient threshold.} We first analyze the setting of gradient threshold $\gamma_{upper}$ and $\gamma_{low}$ by investigating a wide range of potential values. In table~\ref{tab:depth}, we list all threshold we have compared. In fact, different gradient threshold may influence efficiency and effectiveness. In Fig.~\ref{fig:threshold}, we show the performance and efficiency of each setting. With the increment of $\gamma_{low}$, our model deal more moderate samples at an early stage, which accelerate speed but bring significant performance drop. A similar situation also occur when we increase the value of $\gamma_{upper}$. As the growth of $\gamma_{upper}$, the proposed model exhibit promising efficiency with degradation of performance. Since the middle or first stage is unable to deal severe samples well, we think too low $\gamma_{upper}$ and $\gamma_{low}$ may bring obvious performance drop. However, as illustrated in Fig.~\ref{fig:threshold}, the model becomes slower with a decrease of $\gamma$. To achieve a balance between efficiency and performance, we adopt `Our\_YU2L2' as default gradient threshold.

\textbf{Depth of different stage.} In this component, we compare the depth setting of each stage. In other words, we adjust the depth of $\Theta_l$ and $\Theta_m$ to verify our settings. In table.~\ref{tab:stage_depth}, we use different depth setting in the early and middle stage for comparison. As shown in table.~\ref{tab:stage_depth}, with the increase of $\Theta_l$, the model show sight PSNR promotion with slower efficiency. Since the early stage is adapted to handle mild patches only, we think too much parameter is meaningless for further promotion. In contrast, the middle stage $\Theta_m$ is utilized to deal with the moderate sample, which carries some texture and details. Therefore, the performance becomes worse when we reduce parameter of $\Theta_m$. Thus, we use a light-weight setting at an early stage and increase the parameters of the middle stage to exhibit an efficient framework. 

\begin{table}
\centering
\begin{tabular}{llllllll} 
\toprule
 & $\Theta_{l}^1$ & $\Theta_{l}^2$ & $\Theta_{l}^3$ & $\Theta_{m}^1$ & $\Theta_{m}^2$ & $\Theta_{m}^3$ &    \\ 
\cline{2-8}
 PSNR &       27.12      & 27.13 & 27.15 & 27.05 & 27.12 & 27.15 &    \\
 Time &       1.81           & 1.91 & 2.21 & 1.48 & 1.81 & 1.99 &     \\
\bottomrule
\end{tabular}
\caption{Comparison of different depth toward early and middle stage on BSDS100. The superscripts of $\Theta$ mean different depth of each stage. The subscripts of $\Theta$ indicate different stage.}
\label{tab:stage_depth}
\end{table}

\begin{table}[]
\begin{tabular}{|l|l|l|}
\hline
     & o Rolling &  Rolling \\ \hline
PSNR & 27.03       & 27.12      \\ \hline
\end{tabular}
\caption{Comparison of with and without rolling strategy on BSDS100.}
\label{tab:rolling}
\end{table}

\textbf{Rolling strategy.} In order to show the effectiveness of the proposed rolling strategy, we investigate models with and without rolling strategy. In table.~\ref{tab:rolling}, 'o Rolling' means model without rolling strategy and ' Rolling' indicates the model with rolling component. Compared with 'o Rolling', the model with a rolling strategy achieve 0.09 dB improvement. Although our model has an auxiliary parameter, we can use them content-adaptively to assure efficiency. Thus, our model achieves superior performance and maintains competitive efficiency by adopting a rolling strategy.

\subsection{Limitations}
As our model achieves a good balance between effectiveness and efficiency, it still exists some limitations. To advance efficiency, we need to crop the image into smaller images and reconstruct them at last. Thus, our model needs additional time to accomplish the reconstruction procedure. The reconstruction cost is far less than the model computational cost, and we have count the reconstruction time into time complexity in efficiency analysis. Besides, the acceleration is influenced by datasets. For instance, our model can accelerate the speed greatly on BSDS100 or DIV2K as the images in BSDS100 or DIV2K have plenty of blank and mild region. Similar acceleration can not occur in General-100 as the images in General-100 are full with texture and edges. However, we think the majority of nature images, which is closed to BSD500 and DIV2K, are occupied with a certain percentage of the blank or mild region. Therefore, our model can perform similar acceleration in real-world scenarios.

\section{Conclusion and further work}
In this article, to address efficiency problem in image SR, we have proposed an end-to-end gradient-aware rolling network. Our model mainly incorporates gradient prior to the image itself and content-adaptively utilize each stage of the deep neural network to super-resolve corrupted images. Moreover, we have proposed a rolling strategy, which super-resolve images with the different set of filters, to resolve frequency conflicts problem. Experiments have shown that our framework not only obtains competitive performance but also achieve appealing efficiency.

There are several directions for us to extend our work. First, we can introduce adversarial loss or perceptual loss in each stage, aiming to restore more realistic details and texture. Second, considering exist framework have to crop the image into patches, we intend to propose a more general framework, which can content-adaptively process different region with the different stride of convolution operation, to boost efficiency.

\ifCLASSOPTIONcaptionsoff
  \newpage
\fi

\bibliographystyle{plain}
\bibliography{egbib}

\end{document}